%% file: sample-manuscript.tex
\documentclass[manuscript,screen]{acmart}

\usepackage{geometry}
\usepackage{todonotes}
\usepackage{setspace}
\usepackage{caption}
\usepackage{graphicx}
\usepackage{url}
\usepackage{multirow}
\usepackage[T1]{fontenc}
\usepackage{natbib}
\usepackage{listings}
\usepackage[utf8x]{inputenc}

\newcommand{\ME}{\mathcal{E}}
\newcommand{\MA}{\mathcal{A}}

\newcommand{\MD}{\mathcal{V}}

\newcommand{\MV}{\mathcal{V}}
\newcommand{\MX}{\mathcal{X}}
\newcommand{\MH}{\mathcal{H}}
\newcommand{\ML}{\mathcal{L}}
\newcommand{\MK}{\mathcal{K}}

\newcommand{\RR}{\mathbb{R}}

\newcommand{\mysum}{\displaystyle \sum_{i=1}^n}
\begin{document}

\title{Exploring LLM Features in Predictive Process Monitoring for Small-Scale Event-Logs}

\author{Alessandro Padella}
\email{alessandro.padella@unipd.it}
\orcid{0009-0008-6399-9364}
\affiliation{%
  \institution{Università degli Studi di Padova}
  \city{Padova}
  \country{Italy}
}

\author{Massimiliano de Leoni}\orcid{0000-0002-8447-5374}
\affiliation{%
  \institution{Università degli Studi di Padova}
  \city{Padova}
  \country{Italy}
}
\email{deleoni@math.unipd.it}

\author{Marlon Dumas}
\orcid{0000-0002-9247-7476}
\affiliation{%
  \institution{University of Tartu}
  \city{Tartu}
  \country{Estonia}
}
\email{marlon.dumas@ut.ee}

\renewcommand{\shortauthors}{Padella et al.}


\begin{abstract}
Predictive Process Monitoring is a branch of process mining that aims to predict the outcome of an ongoing process. Recently, it leveraged machine-and-deep learning architectures. In this paper, we extend our prior LLM-based Predictive Process Monitoring framework, which was initially focused on total time prediction via prompting. The extension consists of comprehensively evaluating its generality, semantic leverage, and reasoning mechanisms, also across multiple Key Performance Indicators. Empirical evaluations conducted on three distinct event logs and across the Key Performance Indicators of Total Time and Activity Occurrence prediction indicate that, in data-scarce settings with only 100 traces, the LLM surpasses the benchmark methods. Furthermore, the experiments also show that the LLM exploits both its embodied prior knowledge and the internal correlations among training traces. Finally, we examine the reasoning strategies employed by the model, demonstrating that the LLM does not merely replicate existing predictive methods but performs higher-order reasoning to generate the predictions.
\end{abstract}





\maketitle

\section{Introduction}\label{sec:intro}
Predictive Process Monitoring (PPM) is a family of techniques that leverages event logs from business processes to generate predictions about the future states or properties of ongoing process instances~\cite{kimcommuzzi2022}. PPM methods vary depending on the prediction target, which can include times~\cite{survey_remaining_time_verenich}, next activities~\cite{ceravoloss}, or process outcomes~\cite{outcome_oriented_teinemaa}. 

Literature has extensively explored machine-and-deep learning models to enhance prediction quality~\cite{crvl}. However, these models typically require large amounts of data for effective training. When the available event log is limited in size, the applicability of such techniques becomes constrained, reducing the overall potential of PPM. As highlighted in~\cite{zimmermann}, data availability remains one of the most significant challenges faced by researchers and practitioners in this domain. Large Language Models (LLMs) provided an alternative for the application of PPM to data-scarce environments since their embedded knowledge enables robust prediction, while structured prompting preserves trace sequences, dependencies, and attributes without extensive fine-tuning~\cite{gruver2023large,requeima2025llm}.

This paper extends our prior LLM-based PPM work~\cite{10.1007/978-3-032-02929-4_16}. 
Our previous work focused on total time prediction via Gemini\footnote{\url{https://docs.cloud.google.com/vertex-ai/generative-ai/docs/models/gemini/2-5-flash?hl=en}} prompting, which showed to be successful in this context when the training consisted of event logs of limited size, it remained unclear whether the prediction quality would extend to other Key Performance Indicators (KPIs) whose predictions are relevant in different contexts. Our previous work also fell short in a comprehensive analysis of the mechanisms and knowledge used behind to enhance the prediction quality. 
In this paper, we aim to extend the prior work investigating three research questions. 
\begin{enumerate}
    \item \textbf{RQ1:} When trained on event logs with a limited number of traces, do LLMs achieve superior prediction quality on a wide range of metrics, compared to the models available in literature?

    \item \textbf{RQ2:} Do LLMs leverage embodied prior knowledge of the process domains to improve predictions?

    \item \textbf{RQ3:} Do LLMs use an ensemble of different prediction models? If so, how would this take place?
\end{enumerate}
To address RQ1, we extend the experimental scenario to  predicting whether certain activities will occur. All results are statistically validated and benchmarked against state‑of‑the‑art models, including CatBoost\cite{CatBoost} and PGTNet\cite{DBLP:conf/caise/ElyasiAS24}.

To address RQ2, we introduce semantic hashing to enhance the prompt formulation. Specifically, we encode all process-related strings, such as trace variable and activity names, into hashed representations. We then repeat the experiments using these modified prompts. This procedure enables us to assess whether the LLM’s performance relies solely on event correlations or also leverages contextual information.

To answer RQ3, we distill the reasoning processes of the LLM into interpretable $\beta$‑learners derived through pattern analysis of 150 traces. These learners are subsequently re‑implemented and evaluated to determine whether they can match or exceed the performance of the original LLM. Additionally, we conduct statistical significance tests to ensure the robustness and consistency of the derived $\beta$‑learners.

The remainder of this paper is organized as follows: Section~\ref{sec:rw} reviews literature related to LLM in the field of process mining. Section~\ref{sec:prel} presents the preliminaries needed to define the approach in Section~\ref{sec:framework}, in which the prompting technique is reported and detailed. Finally, Section~\ref{sec:exps} reports the extended version of experiments, while Section~\ref{sec:concl} concludes the paper.

\section{Related Works}\label{sec:rw}

LLMs are recently attracting growing focus in business process management as well~\cite{10.1007/978-3-031-82225-4_45}. They have proven to be significantly useful for many process mining tasks, such as process modeling~\cite{10.1007/978-3-031-61007-3_18,Berti2025}, log extraction~\cite{DBLP:conf/coopis/DaniDLBBWR24}, anomaly detection~\cite{VANDERAA2021101824}, and they have also been used for assessing the validity of some new given traces for a given process~\cite{DBLP:conf/icpm/RebmannSGA24}.

Lashkevich et al. in~\cite{DBLP:conf/bpm/LashkevichMAD24} provide a state-of-the-art approach that leverages LLMs for enhancing the optimization of waiting times and relies on user-prompted feedback for recommending more effective re-design options. Rebmann et al.~\cite{DBLP:conf/icpm/RebmannSGA24} present an approach for extracting knowledge from textual data, providing textual and synthetically generated benchmark datasets for extracting event logs to assess missing activities and generate them. The work in~\cite{10680620} leverages LLMs to transform textual data into process representations, followed by training a BERT-based deep learning model to predict the next activity in a process.

Berti et al.~\cite{Berti2025} fine-tuned a pretrained LLM using reinforcement learning to generate complete and executable process models from textual descriptions, employing structural and behavioral rewards to improve correctness and reduce invalid generations. Casciani et al.~\cite{CASCIANI2026102642} introduce a retrieval-augmented LLM framework for next activity prediction in predictive process monitoring using past event traces without training, while identifying limits such as interleaving sensitivity and concept drift. The approach in~\cite{10899487} contextually links event logs with additional process-related data from varied sources, enabling LLMs to provide relevant process insights via natural language querying for planning, monitoring, improving operations and anomaly detection~\cite{10.1007/978-3-032-02867-9_19}. Finally, Kubrak et al.~\cite{DBLP:conf/bpm/KubrakBMND24} developed a chatbot-based approach for process analysis where the LLM explains recommendations from a model to enhance explainability. 

\section{Preliminaries}\label{sec:prel}

The starting point for a process mining-based system is an \textit{event log}. An event log is a multiset of \textit{traces}. Each trace is a sequence of events, each describing a particular \textit{process instance} (i.e., a \textit{case}) in terms of the \textit{activities} executed, the associated \textit{timestamps} and other different domain-related \textit{attributes}.

\begin{definition}[Events]
    Let $\mathcal{A}$ be the set of process activities. Let $\mathcal{T}$ be the set of process timestamps. Let $\mathcal{V}=\mathcal{V}_1\times \mathcal{V}_2\times\ldots\times \mathcal{V}_m$ be the Cartesian product of the data attribute sets.
    An event is a tuple $\left(a,t_{start}, t_{end}, \overrightarrow{v}\right) \in\mathcal{A}\times\mathcal{T}^2\times\MV$ where $a$ is the event activity, $t_{start}$ and $t_{end}$ the associated timestamps, and $\overrightarrow{v}$ the vector of associated attributes.
\end{definition}

\noindent
A trace is a sequence of events. The same event can occur in different traces. Namely, attributes may be given the same assignment in different traces. This means that the same trace can appear multiple times, although admittedly under extremely rare conditions, and motivates why an event log has to be defined as a multiset of traces:

\begin{definition}[Traces \& Event Logs]\label{def:event_log}
Let $\mathcal{E}=\mathcal{A}\times\mathcal{T}^2\times\MD$ be the universe of events. A trace $\sigma$ is a sequence of events, i.e. $\sigma \in \mathcal{E}^{*}$.\footnote{The operator * refers to the Kleene star: given a set $A$, $A^*$ contains all the possible finite sequences of elements belonging to $A$.} An event log $\ML$ is a multiset of traces, i.e. $\ML \subset \mathbb{B}\left(\mathcal{E}^{*}\right)$.\footnote{$\mathbb{B}\left(X\right)$ indicates the set of all multisets with the elements in set $X$.}
\end{definition}

\noindent
Given an event $e=\left(a,t_{start},t_{end},\overrightarrow{v}\right)$, the remainder uses the following shortcuts: $ activity\left(e\right)=a$, $start\left(e\right)=t_{start}$, $end\left(e\right)=t_{end}$, $duration\left(e\right)=t_{start}-t_{end}$, $attr\left(e\right)=\overrightarrow{v}$. Also, given a single attribute set $\MV_i$, it is associated an attribute name, i.e. $name(V_i)$ and it can be classified as \textit{global} or \textit{local}, i.e. $type(\MV_i)\in\{global, local\}$ depending on whether the values in it can vary or not in the same trace. We refer to the value of these attributes as $global\left(\sigma\right)=\overrightarrow{g}$ and $local\left(e\right)=\overrightarrow{l}$, and so the equation $global\left(\sigma\right)\displaystyle\oplus local\left(e\right)=attr\left(e\right)$ holds.\footnote{Considering $\displaystyle\oplus$ as the concatenation of vectors e.g.\newline $[1,3,'request\_ created']\oplus[2,True]=[1,3,'request\_ created',2,True]$} Furthermore, given a trace $\sigma=\left\langle e_{1}, \ldots, e_{n}\right\rangle$, $prefix\left(\sigma\right)$ denotes the set of all prefixes of $\sigma$, including $\sigma$, namely $prefix\left(\sigma\right)=\left\{\langle\rangle,\left\langle e_{1}\right\rangle,\left\langle e_{1}, e_{2}\right\rangle, \ldots,\left\langle e_{1}, \ldots, e_{n}\right\rangle\right\}$. 

The goal of a KPI prediction framework is to forecast the KPI value of a process instance that has not completed yet, namely a \textit{running trace}. 

In this paper, the problem is modeled as the estimation of a KPI function $\MK:\mathcal{X}\rightarrow\RR_0$ that given a running trace $\sigma'=\langle e_1,\ldots, e_k\rangle$ eventually completing as $\langle e_{k+1},\ldots,e_n\rangle$, 
returns the KPI value $k\in \RR_0$ after the occurrence of all the events $\langle e_{1},\ldots,e_n\rangle$ in the trace. The input of the KPI function is a set $\mathcal{X}$, since not every approach shares the same encoding for event logs. For instance, in~\cite{10.1007/978-3-031-59465-6_10} the authors encoded traces in an LSTM compatible input, while in~\cite{DBLP:journals/dke/ShoushD25} the traces are encoded in a \textit{comma-separated values} file suitable for a predictor based on a Decision Tree. This requires defining the \textbf{trace-to-instance encoding function} $\rho:\ME^*\rightarrow \mathcal{X}$ with the goal of accurately translating every trace of the event log into an input suitable for the predictive model. This function has proven to be significantly different based on the chosen predictive approach (cf.\ ~\cite{crvl3}).

\begin{figure}[t]
    \centering
    \includegraphics[width=\linewidth]{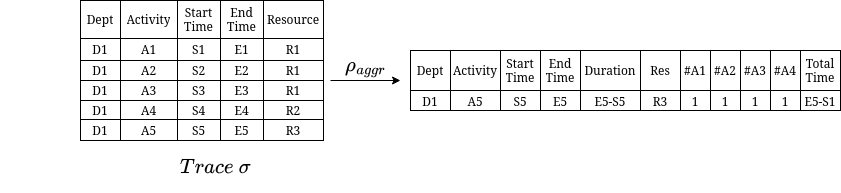}
    \caption{Example of an encoding function applied to a running trace used for predicting the KPI “Total Time”.}
    \Description{Output example of an Aggregated History encoding function, showing how past activities are encoded in new columns.}
    \label{fig:encoding}
\end{figure}

Figure~\ref{fig:encoding} depicts an example of a trace-to-instance encoding function. In it, referred as $\rho_{aggr}$ the trace is preprocessed as a row of a \texttt{.csv} file by adding the past activities in newly generated columns. For each activity in the trace, the number of previous occurrences of that activity is reported in a dedicated column, encoding the number of past executions of the activity. This encoding allows tracking the frequency of all past activities but does not maintain information about their sequential order, recording only the most recent one.

\section{Approach For LLM-based Predictions}\label{sec:framework}

\begin{figure}[t!]
    \centering
     \includegraphics[width=1\linewidth]{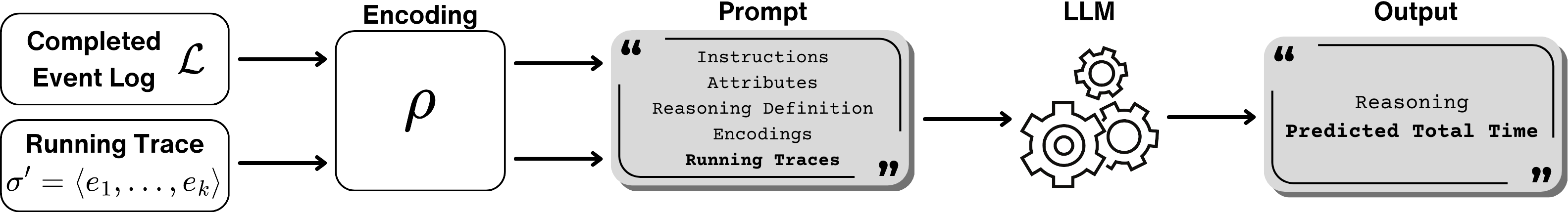}
    \caption{Pipeline outlining the proposed method using LLMs for PPM, in which the KPI is “Total Time”.}
    \Description{Placeholder}
    \label{fig:pipeline}
\end{figure}

This study seeks to leverage the potential of LLMs to develop a framework for PPM, particularly in scenarios where only a small amount of example traces are accessible. Leveraging their embedded knowledge, LLMs can extract and use additional information beyond the event data by incorporating the semantics of events, such as activity names, that traditional models cannot. Figure~\ref{fig:pipeline} depicts the proposed approach. Given an event log of completed traces $\ML$ and a running trace $\sigma'$, a trace-to-instance encoding function $\rho$ is applied to transform them into a structured prompt. This prompt, composed of multiple components, is then used to enable the LLM to estimate the KPI function $\MK$. 

In the remainder of this Section a new trace-to-instance encoding function $\rho_{seq}$ suitable for LLM is introduced in Section~\ref{subsec:framework_enc}, while Section~\ref{subsec:context} defines a context-based prompt suitable for employing an LLM for implementing the KPI function defined in Section~\ref{sec:prel}, while Section~\ref{subsec:explain} provides an example of LLM's output containing the predicted values and its reasoning to achieve it.

\subsection{An Encoding Function for LLMs}\label{subsec:framework_enc}
Exploiting an LLM for developing a PPM framework is a topic that has not yet been explored in process mining (cf.\ Section~\ref{sec:rw}). This section introduces a new LLM-suitable trace-to-instance encoding function $\rho_{seq}:\ME^*\rightarrow \Sigma^*$.\footnote{Considering $\Sigma$ as the set of all finite strings over an alphabet.} 


The input of the $\rho_{seq}$ function is a trace, while the output is a textual prompt that will be later enhanced to become suitable as prompt for an LLM, that will be used as KPI function $\MK$. Specifically, in this case the generic input set of the KPI function $\MX$ is equal to $\Sigma^*$. In $\rho_{seq}$, each trace $\sigma=\langle e_1,\ldots,e_n \rangle$ is mapped into a string composed of three main elements:
\begin{itemize}
\item The values of the global attributes of the trace $global\left(e_n\right)$.
\item A sequence of tuples ${ \left(activity\left(e_i\right),duration\left(e_i\right)\right)\ for\ i=1,\ldots,n}$.
\item The actual value of $\MK(\sigma)$.
\end{itemize}
Formally:
\[
\scalebox{0.9}{$
    \begin{aligned}\rho_{seq}\left(\langle e_1,\ldots,e_n\rangle\right)=& global\left(\langle e_1,\ldots,e_n\rangle\right) \displaystyle\oplus \left(activity\left(e_1\right),duration\left(e_1\right)\right)\displaystyle\oplus\ldots\displaystyle\oplus\notag \\&\left(activity\left(e_n\right),duration\left(e_n\right)\right)\displaystyle\oplus \left(k\right)
    \end{aligned}$}
\]

\noindent
The sets of local attributes have been intentionally excluded, as it has been demonstrated that LLMs are constrained by two primary factors: \textbf{technical} limitations and \textbf{methodological} considerations. From a technical perspective, an LLM can only process a certain number of characters; so it becomes necessary to reduce the size of the input to stay within this maximum quantity, namely the \textbf{Context Length}.\footnote{See \href{https://llm-stats.com/}{https://llm-stats.com/} for an overview of Context Lengths of the latest models.} Note that the Context Length of an LLM is not just a limitation per interaction (e.g., in a chatbot) but an inherent architectural constraint. Additionally, from a methodological standpoint, research demonstrates that the data in an LLM input does not equally impact the model's processing, and the significance of individual data points reduces as the input lengthens, even degrading its performance~\cite{kuratov2024babilong,2024longcontext}. Therefore, we opted to omit local attributes. Conversely, global attributes were retained since they incorporate domain knowledge and have proved to retain more predictive power than local ones (cf.\ Galanti et al.~\cite{DBLP:journals/eaai/GalantiLMNMSM23}).

\begin{figure}
    \centering
    \includegraphics[width=\linewidth]{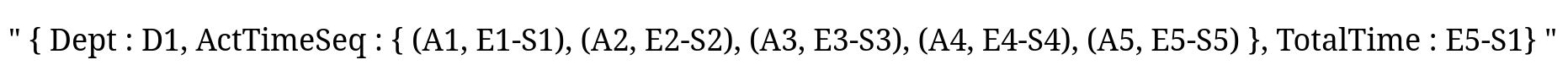}
    \caption{Example of a usage of the Sequential encoding function $\rho_{aggr}$ applied to the same running trace and KPI employed in Figure~\ref{fig:encoding}.}
    \Description{Placeholder}
    \label{fig:seqenc}
\end{figure}
An example of application of the Sequential trace-to-instance encoding function $\rho_{seq}$ is depicted in Figure~\ref{fig:seqenc}, where the example trace reported in Figure~\ref{fig:encoding} has been processed as a string. The result is a string-form Python object primarily composed of three keys: i) \textit{Dept}, associated with the value of the corresponding global attribute; ii) \textit{ActTimeSeq}, associated with the list of tuples where each activity and its duration are recorded; and iii) \textit{TotalTime}, representing the total duration of the trace.

It is important to note that the Sequential is not the only suitable encoding for LLMs. In fact, the proposed approach also supports the \textit{Aggregated History Encoding}, depicted in Figure~\ref{fig:encoding}, as the LLM can predict future values independently of the specific encoding. This flexibility allows the framework to adapt to different encoding strategies, that have been explored in~\cite{10.1007/978-3-032-02929-4_16}.

\subsection{Context-Based Prompting Technique for LLMs in Predictive Process Monitoring}\label{subsec:context}

This Section uses the traces that have been encoded using a trace-to-instance encoding function $\rho$ and incorporates them generating an input suitable for an LLM. In essence we define a prompting technique that allows the model to generate KPI predictions along with corresponding reasoning procedure, starting from the encoded traces. The prompting technique is divided into seven key parts, also reported in the Listing~\ref{lst:train}:  

\begin{itemize}  
    \item \textbf{Initial instruction and Header:} The LLM is introduced to the task with the prompt:  
    \textit{``You are an expert in process mining and machine learning. Your task is to predict the {KPI} of process instances based on event logs of activities, where each process instance is a sequence of activities.''} (Lines 1-2)

    \item \textbf{Attributes and Encoding description:} Contextual information specific to the process is provided, and the trace-to-instance encoding function is described ($\rho_{seq}$ in the example). (Lines 4-12)

    \item \textbf{Output and Reasoning format specification:} The expected structure for predicted values is defined. (Lines 14-22 and 29-35)

    \item \textbf{Running Trace Format Specification:} The format for describing a running trace is specified to the model. (Lines 24-27)

    \item \textbf{Domain-specific background information:} Additional details about the process from which the data have been extracted. (Lines 37-38)

    \item \textbf{Example Data Provision:} Encoded data are provided as example to the model. (Lines 39-45)

    \item \textbf{Running Trace Provision:} The running trace is provided in the same format as the examples, with a custom last activity referred as ``Running'', as described at lines 24-27. (Lines 47-50)
\end{itemize}

Although the proposed encoding is general and applicable to various use cases and KPI, certain information within these seven components must be specified by the process analyst and may be optionally removed. 
\\
\input{prompts/train} 

In Listing~\ref{lst:train} they have been highlighted in bold. They are specifically, (i) the domain-specific background information (Lines 37-38) and (ii) the description of the global attributes (Lines 8-9), as they contain contextual details specific to the process under study.
Notably, these informations are optional and can be excluded if necessary. The remaining sections of the prompt are designed to be generic and can be applied to any event log without modification.

This modularity ensures that the framework can be associated with any encoding function and eventually be customized with details about the single process, while minimizing the effort needed for customization when applying the LLM to different process datasets. 

\subsection{Example of Reasoning to Achieve Predictions}\label{subsec:explain}

The LLM's output is not limited to returning the predicted value: it also reports on the reasoning put in place to conclude the prediction. This is in fact a form of explanation of the predictions, of which one example is given in Listing~\ref{lst:test}.
\\
\input{prompts/test}

Nonetheless, it is a valuable addition to this paper to report and comment on an instance of the prediction's reasoning given as output by the LLM. To do so, Example~\ref{lst:test} reports the output returned by the LLM for Example~\ref{lst:train}. 

The output is returned at the end (see line 32 of Example~\ref{lst:test}). The reasoning of this example is as follows: the LLM has focused on the traces of the provided training log in which the request amount is in the range of 10000 to 25000 (see lines 7-9). This leads to 20 traces (see lines 15-16), for which the median value is computed (see lines 24-25), which is used as predicted value (see explanations given at lines 28 and 29). 

This reasoning certainly provides a valid explanation of the reasoning behind why this prediction is provided. However, this is only an example of the reasoning procedure performed by the LLM, and we observed that the output prompt can vary on the basis of the context and the number of training traces provided. 


\section{Experimental Results}\label{sec:exps}

We assess the effectiveness of our approach in generating valuable predictions using a limited-size training set and compare its performance against state-of-the-art benchmarks across three distinct use cases on two different KPIs. To ensure a robust evaluation, we repeatedly sampled a limited number of traces, trained both a benchmark model and the LLMs on these samples, and conducted multiple experimental runs.
The remainder of this section is organized as follows:
Section~\ref{subsec:traintest} reports how the logs have been divided into training and test sets for use in the experimental setup. The use cases are presented in Section~\ref{subsec:use_cases}, while Section~\ref{subsec:exp_pipeline} outlines the experimental pipeline employed to address the research questions posed in Section~\ref{sec:intro}. Finally, Section~\ref{subsec:results} concludes with the results and their associated analysis.

\subsection{LLM used and Train and Test Split}\label{subsec:traintest}

The whole approach has been implemented in Python and the code is publicly available.\footnote{\url{https://github.com/Pado123/gui_xrecs_presc_analytics}}. The prediction function development has been carried on using \textbf{Gemini 2.5 Flash Thinking}: a state-of-the-art LLM developed by Google DeepMind~\footnote{\url{https://gemini.google.com/app}}, but any choice of LLM is valid. The model is built on a multimodal architecture designed for advanced natural language understanding and generation. As the development of LLMs progresses, we anticipate that the performances of upcoming models using our method may see fast improvements.

Consistently with standard supervised learning practices, we divided the event log $\ML$ into training and test, $\ML^{comp}$ and $\ML^{run}$, respectively. To extract the training log we compute the earliest time $t_{split}$ such that $80\%$ of the identifiers related to traces of $\ML$ are completed. This allows us to define $\ML^{comp}$ as the set of traces of $\ML$ completed at time $t_{split}$, and consequently, define $\ML^{run}$ as $\ML \setminus \ML^{comp}$. The traces of the test log $\ML^{run}$ are truncated to a set $\ML^{trunc}$, namely the set of prefixes, that is obtained from $\ML^{run}$ by removing every event with a timestamp larger than $t_{split}$: $\ML^{trunc}$ only contains the events that occurred before time $t_{split}$. This procedure tries to mimic the reality at time $t_{split}$ and it is in line with the principles introduced in~\cite{DBLP:conf/bpm/WeytjensW21a}. The system is trained on $\ML_{comp}$, the predictions are produced for $\ML_{trunc}$ and tested using its completed form, $\ML_{run}$. Furthermore, to ensure a robust generalization across various process instances and a more balanced comparison with the LLM, we randomly picked 10\% of traces and used them as a validation set $\ML_{valid}$ to apply a Cross-Validation approach to optimize the following parameters of the benchmarks. This technique is a widespread technique in PPM works~\cite{DBLP:conf/coopis/LeoniP24,10.1007/s10618-025-01117-3,PEEPERKORN2024102330} 

It is also worthwhile pointing out that in this paper we use the term \textit{training} to refer to both LLMs and machine- and deep-learning benchmark models, to keep the discussion simple. However, we acknowledge that LLMs are already pre-trained: traces are provided to the LLMs as background, and are not formally used to train its internal parameters.

\subsection{Use Cases and Benchmarks}\label{subsec:use_cases}

The evaluation considers three real-life event logs that are commonly used in PPM:

\begin{itemize}

\item \textbf{Bpi12}\enspace This process has been used by the BPI challenge in 2012\footnote{https://doi.org/10.4121/uuid:3926db30-f712-4394-aebc-75976070e91f}, it contains 8,616 traces, 6 different activities and 1 global attribute: \textit{Requested\_Amount}.

\item \textbf{Bac}\enspace A process referring to a process of a Bank Institution that deals with the closures of bank accounts. It contains 32,429 completed traces, 15 different activities and 2 global attributes: \textit{Closure\_Type}, and \textit{Closure\_Reason}.\footnote{https://github.com/IBM/processmining/tree/main/Datasets\_usecases}

\item \textbf{Hospital}\enspace This process has been provided by an hospital emergency department. The log is made of 37,945 completed traces, contains 46 different activities and 3 global attributes, that are \textit{Triage\_Color},\textit{Triage\_Access} and \textit{Patient\_Age}. Due to confidentiality, this event log cannot be published.

\end{itemize}

For each use case, two prediction tasks, corresponding to two different KPIs:

\begin{itemize}

\item \textbf{Total Time}\enspace of an ongoing case, i.e., a regression task.

\item \textbf{Activity Occurrence} for a selected target class, i.e. a classification task. The chosen 

\end{itemize}

\begin{table}[t]
    \centering
    \begin{tabular}{|c|c|c|c|}\hline
        \textbf{Use Case} & \textbf{Training Traces}  & \textbf{100/Training\_Traces}  & \textbf{Target Activity} \\\hline
         Bpi12&  6792 & 1.45\%&  W\_Nabellen incomplete dossiers\\\hline
         Bac & 25843 & 0.38\%& Service closure Request with BO responsibility\\\hline
         Hospital &  30394&  0.32\%& LABORATORIO\\ \hline
    \end{tabular}
    \caption{Overview of the analyzed use cases, including the number of available training traces, the relative proportion represented by 100 traces, and the target activity selected for the KPI Activity Occurrence.}
    \label{tab:use_cases}
\end{table}

Table~\ref{tab:use_cases} summarizes the analyzed use cases, reporting the proportion that 100 traces represent relative to the total number of training traces. The last column identifies the target activity for the KPI prediction task of Activity Occurrence. These target activities were selected because their execution typically leads to higher time and cost demands. In the Bpi12 use case, the target activity is \texttt{W\_Nabellen incomplete dossiers}, which translates to “Follow up on incomplete files.” This activity occurs when an error in the loan application process requires rework by both the sides of the bank and the customer. Similarly, the activities “Service Closure with BO Responsibility” and “Laboratorio”, that indicates the requirements of further blood analysis, implying additional laboratory processing and resource consumption.

\noindent

For the purpose of setting a benchmark for prediction, we employed two different State-of-the-art techniques for the different KPIs of interest:

\begin{itemize}

\item For Total Time we employed PGTNet architecture from~\cite{DBLP:conf/caise/ElyasiAS24}, that has proven to outperform every framework.

\item For the Activity Occurrence, we employed Catboost from~\cite{CatBoost}, a state-of-the-art model predictor based on machine learning on decision trees, which has been shown to surpass existing prediction frameworks~\cite{galanti2020explainable}.

\end{itemize}

\subsection{Experimental Pipeline}\label{subsec:exp_pipeline}

\label{subsec:methodology}

With the extent of answering to the three research questions posed in Section~\ref{sec:intro}, the experimental pipeline comprises three different main steps:



\subsubsection{Prediction Quality Comparison}\label{subsubsec:preds}

This first step of the pipeline extends the experimental part of~\cite{10.1007/978-3-032-02929-4_16} and addresses RQ1: \textit{When trained on event logs with a limited number of traces, do LLMs achieve superior prediction quality on a wide range of metrics, compared to the models available in literature?}

We measured the accuracy of the LLM when trained on 100 running traces sampled from $\ML^{train}$, that are a limited amount of training traces with respect to the model being trained on the whole training set (cf.\ Table~\ref{tab:use_cases}). 
To evaluate the performance of the prediction models, two key metrics are employed depending on the KPI of interest:

\begin{itemize}
\item \textbf{Mean Absolute Error (MAE)} \enspace
For Total Time prediction, the Mean Absolute Error is used:
\[
    MAE = \frac{1}{n}\mysum |y_i - \hat{y}_i|
\]

where $y_i$ is the actual Total Time, reported in minutes, $\hat{y}_i$ is the predicted value, and $n$ is the number of samples. Lower MAE values indicate better predictive performance.

\item \textbf{F1-Score} \enspace
For activity occurrence prediction, the F1-Score is used:

$$\text{F1-Score} = 2 \cdot \frac{\text{Precision} \cdot \text{Recall}}{\text{Precision} + \text{Recall}}$$

where $\text{Precision} = \frac{TP}{TP + FP}$ and $\text{Recall} = \frac{TP}{TP + FN}$, with $TP$, $FP$, and $FN$ denoting true positives, false positives, and false negatives.

\end{itemize}

Since this 100 traces have been randomly sampled for each run, we repeated the experiments 20 times in order to asses statistical validity, this motivates why the resulting values of MAE/F1-Score are reported associated with the corresponding standard deviation.

\subsubsection{Hashing of the Input Prompts}\label{subsubsec:hashing}
The second step of the experimental pipeline addresses RQ2: \textit{Do LLMs leverage embodied prior knowledge of the process domains to improve predictions?}

To isolate semantic understanding from distributional patterns, we anonymize all context-sensitive strings that could enable the LLM to exploit domain knowledge. Traditional machine-and-deep learning methods remain unaffected by string anonymization since they operate on numerical encodings, conversely LLMs with embodied knowledge may leverage activity semantics. For instance, the activity name “LABORATORIO” implyies laboratory analysis, but also and attribute names as, “Triage\_Color” suggests emergency care. This leads to the fact that we have to anonymize not only the values of the categorical attributes and activities, but also the name of that attributes since they provide information about the process' context.

We define the context-sensitive set $\MH$ as:
\begin{equation}
\MH = \MA \cup \left( \bigcup_{i=1}^n \{v \in \MV_i \mid \text{type}(\MV_i)=\text{global}\} \right) \cup \left( \bigcup_{i=1}^n \{\text{name}(\MV_i) \mid \text{type}(\MV_i)=\text{global}\} \right)
\end{equation}

A deterministic hash function $H: \MH \to \Sigma^4$ maps each $s \in \MH$ to a unique 4-character identifier ($c_1c_2c_3c_4$, $c_i \in \Sigma = {\text{A-Z,0-9}}$), preserving correlations while eliminating semantics. 

In this experimental pipline's phase, after the prompt generation as outlined in Section~\ref{sec:framework}, every $s \in \MH$ is replaced by $H(s)$. Then, the prediction quality in terms of MAE or F1-Score is compared between hashed (anonymized) and non-hashed prompts to quantify the semantic dependency of the LLM. Significant degradation under hashing indicates that the LLM is relying on its embodied knowledge.

Furthermore, we aim to assess the statistical significance between hashed and non-hashed prompts. To this extent we applied the Nemenyi post-hoc test across repeated experiments:

The Nemenyi test, from Nemenyi~\cite{nemenyi1963}, is a global statistical test (such as the Friedman test) that serves to investigate if it is possible to rejected the null hypothesis that the performance of the comparisons on the groups of data is similar. The test makes pair-wise tests of performance and we apply it under the following hypothesis: 

\begin{itemize}

    \item Null hypothesis ($H_0$): No significant difference exists between prediction performance (MAE/F1-Score) of hashed and non-hashed prompts across use cases.

    \item Alternative hypothesis ($H_1$): Significant performance difference exists, with non-hashed prompts expected to outperform due to semantic exploitation.
    
\end{itemize}

\subsubsection{Derivation and Re-implementation of LLM's $\beta$-learners}\label{subsubsec:beta_learners}
The third step of our experimental pipeline addresses RQ3: \textit{Do LLMs use an ensemble of different prediction models? If so, how would this take place?}

To investigate this question, we manually analyzed 50 LLM reasoning explanations per use case and KPI, yielding a total of 150 reasoning traces per KPI. Following the visualization methodology proposed in \cite{zhou2025landscapethoughtsvisualizingreasoning}, we systematically cataloged recurring decision patterns extracted from these textual explanations. Specifically, the aim of this part is to identify how the LLM justifies its predictions as requested in the prompt reported in the Listing~\ref{lst:train}.

These recurring patterns were abstracted into families, the so-called $\beta$-learners. In this context a $\beta$-learners constitutes a mathematically defined and reproducible model that explicitly performs the reasoning strategies that the LLM documented . 

For each use case, every derived $\beta$-learner is applied to the same test set $\ML^{test}$ introduced in~\ref{subsec:traintest}. Then, we compared prediction quality in terms of MAE/F1-Score against the LLM results. 

Furthermore, the statistical significance of performance differences is assessed via Wilcoxon signed-rank test ($\alpha = 0.05$) with the following hypotheses for each $\beta$-learner vs. LLM comparison with the following hypothesis:

\begin{itemize}
\item \textbf{Null hypothesis ($H_0$)}: No significant difference exists between prediction performance (MAE/F1-Score) of the current $\beta$-learner and the one obtained by the LLM.

\item \textbf{Alternative hypothesis ($H_1$)}: A significant difference exists between prediction performance (MAE/F1-Score) of the current $\beta$-learner and the one obtained by the LLM.
\end{itemize}

Furthermore, to determine whether the LLM's predictions merely replicate $\beta$-learner strategies or exhibit more sophisticated inference, we compared each $\beta$-learner's prediction quality \textbf{as employed within the LLM's own reasoning} against the LLM's overall results. In a nutshell, we report the percentage difference in terms of MAE/F1-Score between the LLM and each $\beta$-learner when the LLM has been classified using that $\beta$-learner.

Finally, we want to understand if 150 LLM reasoning per KPI were enough to derive all the $\beta$-learners. To this extent, we employ \emph{Good-Turing frequency estimation}. This statistical technique, originally developed for natural language processing and machine translation, provides a principled framework for estimating the probability of a new unseen event based on the frequency distribution of observed ones~\cite{good1953population}.

The Good-Turing formula estimates the adjusted probability of an item that has been observed $r$ times as:

\begin{equation}
P^*(r) = \frac{(r+1) \cdot N_{r+1}}{N_r \cdot N}
\label{eq:good_turing}
\end{equation}

where:
\begin{itemize}
\item $N_r$ denotes the number of $\beta$-learner patterns that have been observed exactly $r$ times in the training set of 50 derived patterns,
\item $N_{r+1}$ denotes the number of patterns observed exactly $(r+1)$ times,
\item $N$ is the total number of pattern occurrences,
\item $P^*(r)$ is the smoothed probability estimate.
\end{itemize}

The probability of observing a \emph{new} (never before seen) $\beta$-learner is estimated as:

\begin{equation}
P_0 = \frac{N_1}{N}
\label{eq:good_turing_unseen}
\end{equation}

where $N_1$ is the number of $\beta$-learner patterns observed exactly once in the training set.

This approach allows us to compute the expected number of novel $\beta$-learners $E[\text{novel}]$ that would be encountered when applying the method to $m$ new prediction instances (where $m \in \{1, 10, 100\}$):

\begin{equation}
E[\text{novel}] = m \cdot P_0 = m \cdot \frac{N_1}{N}
\label{eq:expected_novel}
\end{equation}

\subsection{Results}\label{subsec:results}

\subsubsection{Quality Prediction and Hashing Results}

\begin{table}[t]
\centering
\caption{MAE (minutes) for hashed/non-hashed prompts}
\begin{tabular}{|l|l|l|l|l|}
\hline
\textbf{Use Case} & \textbf{Model} & \textbf{hash} & \textbf{all\_df} & \textbf{100} $\pm$ \\
\hline
\multirow{4}{*}{Bpi12}
& CatBoost Regressor & - & $6846 $ & $9394 \pm 1275$ \\
& PGTNet & - & $3888 $ & $8856 \pm 576$ \\
& 2.5 flash & no &- & $6508 \pm 235$ \\
& 2.5 flash & yes &- & $9246 \pm 873$ \\
\hline
\multirow{4}{*}{Bac}
& CatBoost Regressor & - & $2647 $ & $6393 \pm 365$ \\
& PGTNet & - & $1245 $ & $4753 \pm 472 $ \\
& 2.5 flash & no &- & $2265 \pm 1072$ \\
& 2.5 flash & yes &- & $3880 \pm 3254$ \\
\hline
\multirow{4}{*}{hospital}
& CatBoost Regressor & - &253 & $259 \pm 52$ \\
& PGTNet & - &97 & $132 \pm 32$ \\
& 2.5 flash & no &- & $115 \pm 34$ \\
& 2.5 flash & yes &- & $2077 \pm 232$ \\
\hline
\end{tabular}\label{tab:mae_hash}
\end{table}

Table~\ref{tab:mae_hash} presents MAE in minutes for the prediction of the KPI Total Time across Bpi12, Bac, and Hospital use cases, comparing with the benchmark of CatBoost Regressor and PGTNet, as introduced in Section~\ref{tab:use_cases}. Results also reports the MAE when the benchmarks used all training data (all\_df), scarce-data (100 traces, \~1-0.3\% of logs per Table~\ref{tab:use_cases}), and hashed vs. non-hashed, associated with the standard deviation.

When trained on 100 traces, non-hashed LLM yields competitive MAE: Bpi12 ($6508 \pm 235$) surpasses CatBoost ($9394 \pm 1275$) and nears PGTNet ($8856 \pm 576$) while Hospital ($115 \pm 34$) outperforms both ($259 \pm 52$, $132 \pm 32$). In a similar manner Bac ($2265 \pm 1072$) outperforms both PGTNet ($4753 \pm 472$) and CatBoost.

Hashing activity/attribute semantics (e.g., masking "LABORATORIO") degrades LLM's performance of Bpi12 to $9246 \pm 873$ (+42\%), Hospital to $2077 \pm 232$ (+1700\%), Bac to $3880 \pm 3254$.

\begin{table}[t]
\centering
\caption{F1-Score for hashed/non-hashed prompts}
\begin{tabular}{|l|l|l|l|l|}
\hline
\textbf{Use Case} & \textbf{Model} & \textbf{hash} & \textbf{all\_df} & \textbf{100} $\pm$ \\
\hline
\multirow{3}{*}{Bpi12}& CatBoost Classifier& - & 0.80& 0.72 $\pm$ 0.07\\
& 2.5 flash & no &-& 0.77 $\pm$ 0.06\\
& 2.5 flash & yes &-& 0.74 $\pm$ 0.06\\
\hline
\multirow{3}{*}{Bac}& CatBoost Classifier& - & 0.95& 0.78 $\pm$ 0.08\\
& 2.5 flash & no &-& 0.98 $\pm$ 0.04\\
& 2.5 flash & yes &-& 0.96 $\pm$ 0.04\\
\hline
\multirow{3}{*}{hospital}& CatBoost Classifier& - &0.90& 0.90 $\pm$ 0.01\\
& 2.5 flash & no &-& 0.90 $\pm$ 0.08\\
& 2.5 flash & yes &-& 0.84 $\pm$ 0.06\\
\hline
\end{tabular}
\label{tab:f1-score_results}
\end{table}

Table~\ref{tab:f1-score_results} reports F1-scores for predicting targeted activity occurrence, comparing CatBoost Classifier as benchmark. As for Table~\ref{tab:mae_hash}, the results for the benchmark have been also reported when it has been trained on the whole $\ML^{train}$

When trained on 100 traces, non-hashed LLM achieves comparable or superior performance with respect to the benchmark: Bpi12 (0.77 $\pm$ 0.06 vs. 0.72 $\pm$ 0.07), Bac (0.98 $\pm$ 0.04 markedly exceeding 0.78 $\pm$ 0.08), Hospital (0.90 $\pm$ 0.08 matching 0.90 $\pm$ 0.01). Furthermore, low standard deviations confirm prediction stability, demonstrating LLMs' capacity to forecast activity occurrences despite minimal training traces. Addressing RQ1.

Hashing yields consistent yet moderate degradation, such as Bpi12 to 0.74 $\pm$ 0.06 (-4\%), Bac to 0.96 $\pm$ 0.04 (-2\%), Hospital to 0.84 $\pm$ 0.06 (-7\%).

\begin{table}[t!]
\centering
\caption{Summary of Nemenyi Post-Hoc Test Results}
\begin{tabular}{l l c c c}
\hline
\textbf{Task} & \textbf{Use Case} & \textbf{Difference} & \textbf{P-value} & \textbf{Significance} \\
\hline
\multirow{3}{*}{Total Time} & Bpi12 & 42.06\% & 0.002 & ** \\
& Bac & 71.25\% & $<$0.001 & *** \\
& Hospital & 1702\% & 0.002 & ** \\
\hline
\multirow{3}{*}{Activity Occurrence} & Bpi12 & 0.03 & 0.018 & * \\
& Bac & 0.02 & 0.040 & * \\
& Hospital & 0.06 & $<$0.001 & *** \\
\hline
\end{tabular}
\label{tab:nemenyi_summary}
\end{table}

Table~\ref{tab:nemenyi_summary} outlines the results on the Nemenyi tests for addressing RQ2. All the comparisons reject the null hypothesis of no difference between hashed and non-hashed prompts, confirming the LLM's reliance on semantic embodied knowledge. For total time prediction, effect sizes are substantial, also rising to 1702\% difference in Hospital use case. These findings, aligned with MAE/F1 trends in Tables~\ref{tab:mae_hash} and \ref{tab:f1-score_results}, demonstrate that LLMs exploit event semantics. 

\subsubsection{\texorpdfstring{$\beta$}{beta}-Learner Performance Analysis}
For the prediction of the KPI Total Time, the extracted $\beta$-learners are:

\begin{itemize}
\item \textbf{knn act + mean/median/mode}: $k$-nearest neighbors on activity-based representations with different aggregation functions.
\item \textbf{knn att + mean/median/mode}: $k$-nearest neighbors on attribute-based representations with analogous aggregations functions.
\item \textbf{time seq + mean/median/mode}: models that exploit the temporal sequence of events with different aggregation functions.
\item \textbf{path pred + mean/median/mode}: models based on predicted future paths of a case, again with different aggregations functions.
\end{itemize}

For prediction of the KPI activity occurrence prediction, different $\beta$-learners have been identified:

\begin{itemize}
\item \textbf{Activity-Based}: Based on analyzing the sequence of activities
\item \textbf{State-Based}: Based on analyzing the last event
\item \textbf{Att-Based}: Based on analyzing the trace attributes values.
\item \textbf{Positive Evidence}: Checking if the activity to prediction already occurred or not.
\end{itemize}
The details on how the single $\beta$-learners have been implemented are reported in the code url.

\begin{table}[t!]
\centering
\resizebox{.9\textwidth}{!}{\begin{tabular}{|l|l|l|l|c|c|c|c|}
\hline
\textbf{Use Case} & \textbf{Model} & \textbf{all\_df} & \textbf{\textbf{100} $\pm$} & \textbf{\#best50 no} & \textbf{Significance}  & \textbf{Occurrence}  & $\Delta$\textbf{LLM when occurring}\\
\hline
\multirow{13}{*}{Bpi12}
& 2.5 flash no-hash & -  & $6508 \pm 235$ & 39 &-  &- &-\\
& knn act mean  & 8356 & $8651 \pm 393$ & 2 & ***  &5 & 6\%\\
& knn act median  & 8082 & $8780 \pm 524$ & 2 & ***  &5 & 8\%\\
& knn act mode  & 13463 & $13200 \pm 1002$ & 0 & ***  &6 & 9\% \\
& knn att mean  & 13261 & $8974 \pm 245$ & 1 & ***  &4 & 75\%\\
& knn att median  & 14253 & $8957 \pm 453$ & 0 & ***  &2 & 51\%\\
& knn att mode  & 14591 & $16077 \pm 1690$ & 0 & ***  &4 & 24\%\\
& time seq mean  & 8764 & $9127 \pm 537$ & 1 & ***  &3 & 26\%\\
& time seq median  & 8016 & $8614 \pm 772$ & 4 & ***  &6 & 7\%\\
& time seq mode  & 12433 & $12710 \pm 1636$ & 0 & ***  &5 & 9\%\\
& path pred mean  & 15023 & $15060 \pm 3087$ & 0 & ***  &4 & 23\%\\
& path pred median  & 12349 & $13468 \pm 2819$ & 1 & ***  &4 & 23\%\\
& path pred mode  & 8546 & $15998 \pm 4578$ & 0 & ***  &6 & 11\%\\
\hline

\multirow{13}{*}{Bac}

& 2.5 no-hash & -  & $2265 \pm 874$ & 36 &-  &- &-\\
& knn act mean  & 3994 & $6788 \pm 3972$ & 1 & *** &4 & 24\%\\
& knn act median  & 3563 & $3095 \pm 437$ & 4 & **  &9 & 10\%\\
& knn act mode  & 3372 & $5503 \pm 3434$ & 0 & **  &6 & 22\%\\
& knn att mean  & 21296 & $8094 \pm 2874$ & 0 & ***  &3 & 41\%\\
& knn att median  & 5401 & $4678 \pm 1969$ & 1 & **  &3 & 22\%\\
& knn att mode  & 5379 & $14964 \pm 879$ & 1 & ***  &3 & 11\%\\
& time seq mean  & 6917 & $9224 \pm 3820$ & 0 & ***  &2 & 8\%\\
& time seq median  & 6966 & $3916 \pm 531$ & 1 & ***  &5 & 12\%\\
& time seq mode  & 8702 & $9534 \pm 4135$ & 0 & ***  &1 & 22\%\\
& path pred mean  & 54692 & $6843 \pm 3820$ & 0 & ***  &6 & 61\% \\
& path pred median  & 3569 & $3249 \pm 445$ & 4 & ***  &9 & 8\%\\
& path pred mode  & 190092 & $5927 \pm 2699$ & 0 & ***  &4 & 6\%\\
\hline

\multirow{13}{*}{Hospital}
& 2.5 flash no-hash & -  & $115 \pm 60$ & 45 &-  &- &-\\
& knn act mean  & 167 & $442 \pm 379$ & 0 & **  &4 & 24\%\\
& knn act median  & 388 & $175 \pm 15$ & 1 & **  &7 & 11\%\\
& knn act mode  & 458 & $503 \pm 653$ & 0 & ***  &2 & 12\%\\
& knn att mean  & 617 & $566 \pm 437$ & 0 & ***  &3 & 12\%\\
& knn att median  & 189 & $183 \pm 27$ & 1 & ***  &5 & 80\%\\
& knn att mode  & 1479 & $626 \pm 546$ & 0 & ***  &7 & 68\%\\
& time seq mean  & 661 & $670 \pm 805$ & 0 & ***  &4 & 11 \%\\
& time seq median  & 684 & $287 \pm 677$ & 0 & ***  &2 & 61\%\\
& time seq mode  & 643 & $708 \pm 771$ & 1 & ***  &4 & 34\%\\
& path pred mean  & 383 & $382 \pm 320$ & 0 & ***  &4 & 32\%\\
& path pred median  & 159 & $174 \pm 15$ & 2 & **  &4 & 14\%\\
& path pred mode  & 282 & $473 \pm 618$ & 0 & ***  &7 & 31\%\\
\hline
\end{tabular}}
\caption{Complete MAE results across all use cases and models, also paired with the possible outcome of $\beta$-learners.}
\label{tab:mae_results}
\end{table}

Tables \ref{tab:mae_results} and \ref{tab:f1-score_results} present the MAE and F1-Score results describing the $\beta$-learners’ predictive performance compared to the LLM. The tables systematically compare the re-implementations of the LLM’s reasoning patterns, i.e., the $\beta$-learners such as knn-act/att-mean/mode (nearest neighbors on activity/attribute features), time-seq-mean (temporal sequence aggregation), and path-pred-mean/mode (predicted future path estimation), across all three use cases under both the full $\ML^{train}$ and when trained on 100 traces.

Table \ref{tab:mae_results} reports almost all significantly different results (cf. Significance column), showing that the LLM consistently outperforms every associated $\beta$-learner with strong statistical support. This trend is slightly different in Table \ref{tab:f1-score_results}, where, for the Bpi12 use case, the $p$-value indicating the difference is less significant than in the others. However, the results remain stable and statistically meaningful.

In the $\Delta$ LLM column, the percentage difference in MAE and F1-Score relative to the LLM is reported for Table \ref{tab:mae_results} and \ref{tab:f1-score_results}, respectively. Since every result is far from 0, we can conclude that the LLM model does not simply replicate the $\beta$-learners, but instead performs substantially better.

\begin{table}[t!]
\centering
\small
\caption{F1-Score results related to $\beta$-learners.}
\begin{tabular}{|l|l|l|l|l|c|c|c|}
\hline
\textbf{Use Case} & \textbf{Model} & \textbf{hash} & \textbf{all\_df} & \textbf{100} $\pm$ & \textbf{Significance}  & \textbf{Occurrence}  &$\Delta$ LLM when occurring\\
\hline
\multirow{5}{*}{Bpi12}
& 2.5 flash & no & - & $0.77 \pm 0.06$ & -  &- &-\\
& Activity-Based & - & 0.71 & $0.71 \pm 0.05$ & **  &15 & 0.03\\
& State-Based & - & 0.70 & $0.70 \pm 0.02$ & *  &14 & 0.02\\
& Positive Evidence & - & 0.57 & $0.57 \pm 0.01$ & **  &17 & 0.07  \\ 
& Att-Based & - & 0.50 & $0.50 \pm 0.10$ & **  &12 & 0.06 \\
\hline

\multirow{5}{*}{Bac}
& 2.5 flash & no & - & $0.98 \pm 0.04$ & -  &- &-\\
& Activity-Based & - & 0.78 & $0.75 \pm 0.04$ & ***  &10 & 0.04\\
& State-Based & - & 0.72 & $0.72 \pm 0.05$ & ***  &21 & 0.24\\
& Positive Evidence & - & 0.58 & $0.58 \pm 0.04$ & ***  &10 & 0.25\\
& Att-Based & - & 0.52 & $0.52 \pm 0.03$ & ***  &9 & 0.19\\
\hline

\multirow{5}{*}{Hospital}
& 2.5 flash & no & - & $0.90 \pm 0.08$ & -  &- &-\\
& Activity-Based & - & 0.82 & $0.80 \pm 0.05$ & ** &15 & 0.09 \\
& State-Based & - & 0.75 & $0.71 \pm 0.04$ & ***  &10 & 0.18\\
& Positive Evidence & - & 0.58 & $0.58 \pm 0.04$ & ***  &12 & 0.14\\
& Att-Based & - & 0.53 & $0.53 \pm 0.06$ & ***  &14 & 0.25\\
\hline
\end{tabular}
\end{table}

\subsubsection{Good-Turing Results}
\begin{table}[t!]
\centering
\caption{Good-Turing Smoothing Analysis: Novel $\beta$-Learner Discovery Rate}
\centering
\begin{tabular}{|c|c|}
\hline
 \textbf{Task} & \textbf{Expected Novel $\beta$-Learners} \\
\hline
 Total Time & $m=1$: 0.000, $m=10$: 0.001, $m=100$: 0.003 \\\hline
 Activity Occurrence & $m=1$: 0.000, $m=10$: 0.004, $m=100$: 0.014\\ \hline
\end{tabular}
\label{tab:good_turing_all}
\end{table}

\begin{figure}[h!]
\caption{KPI convergences of various use cases.}
\centering
\includegraphics[width=.90\linewidth]{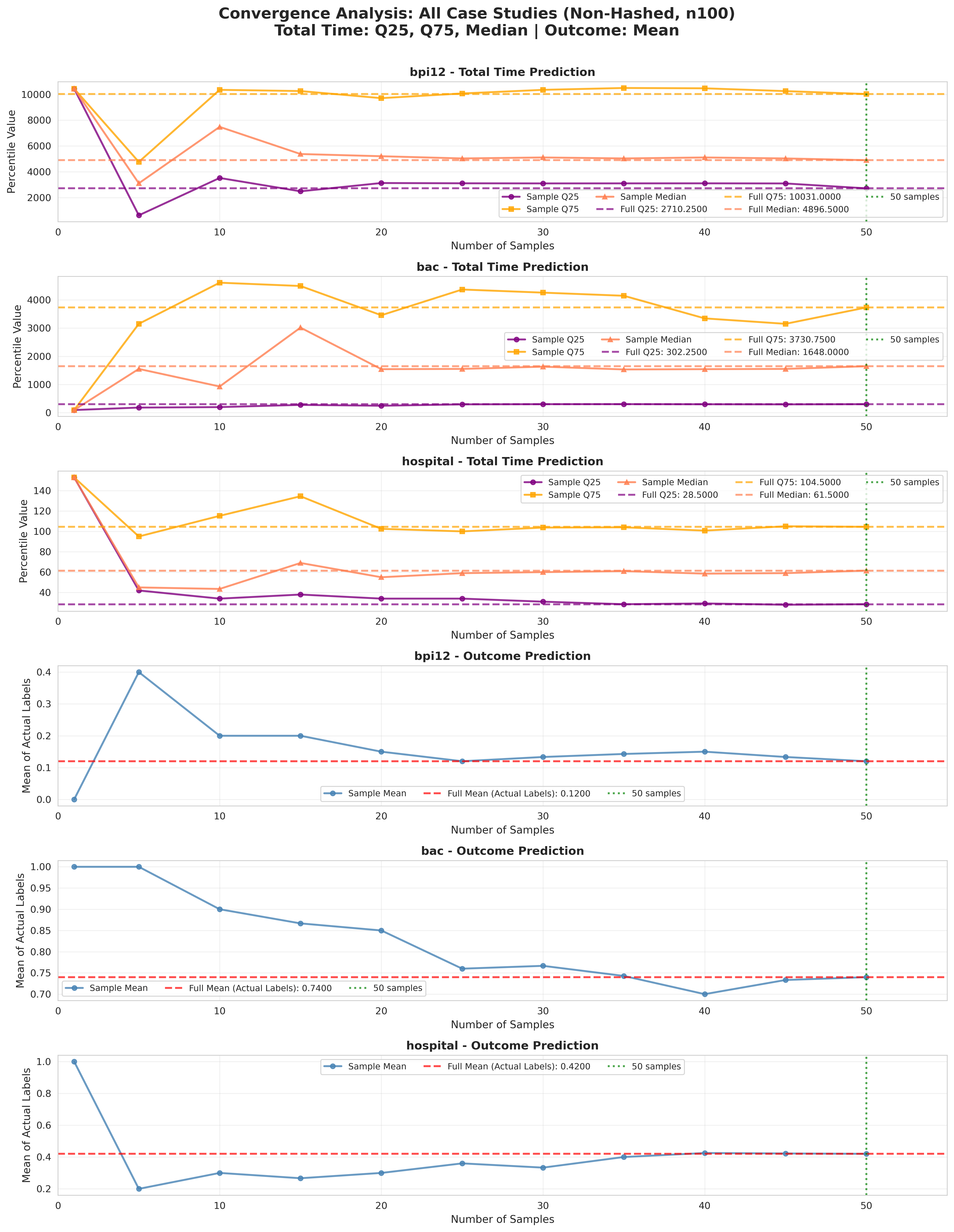}
\label{fig:convergence}
\Description{Placeholder}
\end{figure}

Table~\ref{tab:good_turing_all} provides results in terms of \textit{the probability of encounter a new $\beta$-learner} analyzing other 1, 10 or 100 LLM's outputs. Strong empirical evidence supporting the claim that the probability of discovering novel $\beta$-learner patterns in future predictions approaches zero. Expected novel $\beta$-learners for $m=1,10,100$ additional traces approach zero in every use case and KPI. Specifically, after 1 new running trace, the probaility is exactly 0 for both KPI of Total Time and Activity Occurrence. After 10 new running traces, it raises to 0.1\% and 0.4\% for respectively Total Time and Activity Occurrence, reaching 0.3\% and 1.4\% after 100 new running traces.

Furthermore, Figure~\ref{fig:convergence} plots the convergence curves of the different KPI distributions (Total Time on top rows and Activity Occurrence on bottom rows) when analyzing a given number of traces, also diveded per use case. From the different plot it is possible to visualize that the convergence in the distribution has been already reached after roughly 30 traces.

\section{Conclusion}\label{sec:concl}
This paper advances our previous work~\cite{10.1007/978-3-032-02929-4_16} in the field of Predictive Process Monitoring in data-scarce environments through a comprehensive LLM prompting framework. It rigorously addresses three new research questions via empirical analysis across three public and private event logs and two different Key Performance Indicators (KPIs).

RQ1 confirms the LLM's superiority when the model is trained on 100 traces, outperforming state-of-the-art benchmarks for both KPIs. This demonstrates that the LLM is a viable alternative for implementing Predictive Process Monitoring frameworks.

RQ2 empirically proves semantic leverage: the context in the prompts was hashed and the experiments repeated, showing that the LLM exploits its embodied knowledge to provide predictions for running traces. The results were also statistically verified using the Nemenyi post-hoc test.

RQ3 reveals the reasoning anatomy: manual inspection of 50 traces for each use case and KPI led to the identification of different predictors that the LLM employs, the so-called $\beta$-learners. These have been shown to include all possible ones through Good-Turing validation. This paper demonstrates not only that the LLM outperforms every model it mimics but also that it performs more complex analyses than those models.

Future work will focus on exploring new LLMs, given the rapid evolution of the field, and will include user studies on the reliance on $\beta$-learners. Another direction is to extend this work toward a Prescriptive Process Analytics framework, enabling not only predictions but also actionable recommendations derived from them.
\newpage




\bibliographystyle{ACM-Reference-Format}

\bibliography{bibliography}

\pagebreak

\pagebreak

\end{document}

%% file: prompts/train.tex
\lstset{
basicstyle=\fontsize{6.1}{8}\ttfamily,
    label={lst:train},
    numbers=left,
    caption={Prompting technique example for a loan application process (Bpi12), for the KPI of Total Time. Lines that have to be provided by the process analyst are marked in bold. In the example, only 2 training traces are provided due to space limitation.},
    captionpos=b,
    moredelim=**[is][\bfseries]{@}{@},
    float=t
} 
\begin{lstlisting}
You are an expert in process mining and machine learning. Your task is to predict the 'total time' of 
process instances based on event logs, as each process instance is a sequence of activities.

A event log is a collection of traces, where each trace represents a process instance. 
Each trace is mapped as a sequence of activities and integers representing the minutes since the start
of the process.
The log is represented as a python list containing one dictionary for each trace. Included in it are:
@- the key "AMOUNT_REQ", representing the total amount of euros requested in the loan application.@
- the key "ActTimeSeq", which value is a list of [activity, cumulative elapsed minutes] 
- The key "total_time", which value is the total execution time in minutes from the start of the activity, 
that is the value to predict.

All interactions will be structured in the following way, with the appropriate values filled in.

[[ ## reasoning ## ]]
{your step-by-step reasoning}

[[ ## answer ## ]]
{your predicted total time as an integer}

[[ ## completed ## ]]

In adhering to this structure, your objective is to analyze the event log, and apply reasoning to predict 
the total time for a new case. This case belongs to a not-yet-completed process instance, represented by the 
label "Running" in "ActTimeSeq", indicating that more activities are expected before reaching the conclusion
of the process instance.

Ensure to articulate each step of your thought process in the reasoning field, detailing how you identify
relationships with past cases and leverage your intuition about the meaning of activities to arrive at the 
solution. The answer should be the final prediction of the total time for the given process instance.
Respond with the corresponding output fields, starting with the field [[ ## reasoning ## ]], 
then [[ ## answer ## ]], and then ending with the marker for [[ ## completed ## ]].

Your task is to learn from them and predict the 'total time' values for that traces.

@The process deals with a loan application process from a Dutch financial institution. It has been provided 
in the Business Process Intelligence (BPI) challenge in 2012. @
The following list shows some completed example cases with their total times:

     {"AMOUNT_REQ": 5000.0, "ActTimeSeq": [["W_Completeren aanvraag", 11],
     ["W_Nabellen offertes", 1464], ["W_Nabellen offertes", 7486]], "total_time": "7486"}
     {"AMOUNT_REQ": 15000.0, "ActTimeSeq": [["W_Completeren aanvraag", 13],
     ["W_Nabellen offertes", 14], ["W_Validate application", 4328], ["W_Validate application", 8792]], 
     "total_time": "8792"}
        
Now predict the total time for this new uncompleted case, considering that the case is still running:

    {"Application_1000386745": {"AMOUNT_REQ": 18000.0, 
    "ActTimeSeq": [["W_Completeren aanvraag", 2], ["W_Nabellen offertes", 8571], ["Running"]]}
\end{lstlisting}

%% file: prompts/test.tex
\lstset{basicstyle=\fontsize{6.1}{8}\ttfamily, label={lst:test},
caption={LLM's output and reasoning returned after the prompting technique example for a loan application process.},
captionpos=b, float=tb} 
\begin{lstlisting}
[[ ## reasoning ## ]]
The new case is for an application with a 'AMOUNT_REQ' of 18000.  The process instance 
has executed activities 'W_Completeren aanvraag' and 'W_Nabellen offertes'.
The process is currently 'Running'. 
To predict the total time, we will look at historical cases with similar characteristics.

1. Filter for similar Application Type and Requested Amount:
   We focus on cases with 'AMOUNT_REQ' 
   in a range around 18000.0 (e.g., 10000.0 to 25000.0) to find comparable instances.

2. Identify relevant historical cases:
   From the provided log, we select the cases that match 'AMOUNT_REQ' within the range of 10000.0 to 25000.0. 
   We list their total times:
   10000.0: 9713, 10000.0: 18803, 11500.0: 4771, 12000.0: 18924, 12000.0: 8102,
   13000.0: 6645, 13000.0: 21550, 14000.0: 9713, 15000.0: 8049, 15000.0: 17011

3. Calculate the median total time:
   To get a robust estimate and reduce the impact of outliers, we calculate the median of these total times.
   Sorted total times: 4771, 5299, 5975, 5975, 6645, 6645, 7432, 8049, 8102, 9713, 9713, 
   12991, 13056, 17011, 17434, 18803, 18924, 19826, 21550, 71762.
   71762 is detected as an outlier and it will be removed from the analysis.
   Since there are 19 values, the median is the 10th value,
   which is 9713. Median = 9713.

4. Predict total time:
   We use the median total time of the similar historical cases as our prediction
   for the new running case.

[[ ## answer ## ]]
9713
[[ ## completed ## ]]
\end{lstlisting}